\documentclass{article}

\usepackage{PRIMEarxiv}

\usepackage[utf8]{inputenc} 
\usepackage[T1]{fontenc}    
\usepackage{hyperref}       
\usepackage{url}            
\usepackage{booktabs}       
\usepackage{amsfonts}       
\usepackage{nicefrac}       
\usepackage{microtype}      
\usepackage{lipsum}
\usepackage{fancyhdr}       
\usepackage{graphicx}       
\usepackage{natbib}

\usepackage{amsmath,amsfonts,bm}

\def\eqref#1{equation~\ref{#1}}

\def\1{\bm{1}}

\DeclareMathAlphabet{\mathsfit}{\encodingdefault}{\sfdefault}{m}{sl}
\SetMathAlphabet{\mathsfit}{bold}{\encodingdefault}{\sfdefault}{bx}{n}

\usepackage{caption}
\usepackage{subcaption}

\title{Rethinking Backdoor Data Poisoning Attacks in the Context of Semi-Supervised Learning}

\author{
  Marissa Connor, Vincent Emanuele \\
  Embedded Intelligence \\
  \texttt{\{marissa, vince\}@embedintel.com} \\
}

\begin{document}
\maketitle

\begin{abstract}
Semi-supervised learning methods can train high-accuracy machine learning models with a fraction of the labeled training samples required for traditional supervised learning. Such methods do not typically involve close review of the unlabeled training samples, making them tempting targets for data poisoning attacks. In this paper we investigate the vulnerabilities of semi-supervised learning methods to backdoor data poisoning attacks on the unlabeled samples. We show that simple poisoning attacks that influence the distribution of the poisoned samples' predicted labels are highly effective - achieving an average attack success rate as high as $96.9\%$. We introduce a generalized attack framework targeting semi-supervised learning methods to better understand and exploit their limitations and to motivate future defense strategies. 
\end{abstract}


\section{Introduction}

Machine learning models have achieved high classification accuracy through the use of large, labeled datasets. However, the creation of diverse datasets with supervised labels is time-consuming and costly. In recent years, semi-supervised learning methods have been introduced which train models using a small set of labeled data and a large set of unlabeled data. These models achieve comparable classification accuracy to supervised learning methods while reducing the necessity of human-based labeling. The lack of a detailed human review of training data increases the potential for attacks on the training data. 

Data poisoning attacks adversarially manipulate a small number of training samples in order to shape the performance of the trained network at inference time. Backdoor attacks, one type of data poisoning attack, insert a backdoor (or an alternative classification pathway) into a trained model that can cause sample misclassification through the introduction of a trigger (a visual feature that is added to a poisoned sample)~\citep{gu2017badnets}. We focus our analysis on backdoor attacks which poison the unlabeled data in semi-supervised learning. In this setting, backdoors must be introduced in the absence of training labels associated with the poisoned images. Recent semi-supervised learning methods achieve high accuracy with very few labeled samples~\citep{xie2020unsupervised,berthelot2020remixmatch,sohn2020fixmatch} using the strategies of pseudolabeling and consistency regularization which introduce new considerations when assessing the risk posed by backdoor attacks. Pseudolabeling assigns hard labels to unlabeled samples based on model predictions~\citep{lee2013pseudo} and is responsible for estimating the training labels of unlabeled poisoned samples. Consistency regularization encourages augmented versions of the same sample to have the same network output~\citep{sajjadi2016regularization} and requires attacks to be robust to significant augmentations. 

In this paper we analyze the impact of backdoor data poisoning attacks on semi-supervised learning methods by first reframing the attacks in a setting where pseudolabels are used in lieu of training labels and then highlighting a vulnerability of these methods to attacks which influence expected pseudolabel outputs. We identify characteristics of successful attacks, evaluate how those characteristics can be used to more precisely target semi-supervised learning, and use our insights to suggest new defense strategies. We make the following contributions:
\begin{itemize}
\item We show two simple, black-box backdoor attacks which vary the predicted pseudolabel behavior are highly effective against semi-supervised learning methods.
\item We analyze unique dynamics of data poisoning during semi-supervised training and identify characteristics of attacks that are important for attack success.
\item We introduce a generalized attack framework targeting semi-supervised learning.
\end{itemize}

\section{Background}

\subsection{Data Poisoning}
We focus on integrity attacks in data poisoning which maintain high classification accuracy while encouraging targeted misclassification. Instance-targeted attacks and backdoor attacks are two types of integrity attacks. Instance-targeted attacks aim to cause a misclassification of a specific example at test time~\citep{shafahi2018poison,zhu2019transferable,geiping2020witches,huang2020metapoison,aghakhani2021bullseye}. While an interesting and fruitful area of research, we do not consider instance-targeted attacks in this paper and instead focus on backdoor attacks. Traditional backdoor attacks introduce triggers into poisoned images during training and adapt the images and/or the training labels to encourage the network to ignore the image content of poisoned images and only focus on the trigger~\citep{gu2017badnets,turner2019label, saha2020hidden,zhao2020clean}. They associate the trigger with a specific target label $y_t$.

There are two types of backdoor data poisoning attacks against supervised learning which use different strategies to encourage the creation of a backdoor: dirty label attacks which change the training labels from the ground truth label~\citep{gu2017badnets} and clean label attacks which maintain the ground truth training label while perturbing the training sample in order to increase the difficulty of sample classification using only image-based features~\citep{turner2019label,saha2020hidden,zhao2020clean}. In both of these attacks, the labels are used to firmly fix the desired network output even as the images appear confusing due to perturbations or having a different ground truth class. Greater confusion encourages the network to rely on the triggers, a constant feature in the poisoned samples.

\subsection{Semi-Supervised Learning}
The goal of semi-supervised learning is to utilize unlabeled data to achieve high accuracy models with few labeled samples. This has been a rich research area with a variety of proposed techniques~\citep{van2020survey,yang2021survey}. We focus on a subset of recent semi-supervised learning techniques that have significantly improved classification performance~\citep{xie2020unsupervised,berthelot2020remixmatch,sohn2020fixmatch}. These techniques make use of two popular strategies: consistency regularization and pseudolabeling. Consistency regularization is motivated by the manifold assumption that transformed versions of inputs should not change their class identity. In practice, techniques that employ consistency regularization encourage similar network outputs for augmented inputs~\citep{sajjadi2016regularization,miyato2018virtual,xie2020unsupervised} and often use strong augmentations that significantly change the appearance of inputs. Pseudolabeling uses model predictions to estimate training labels for unlabeled samples~\citep{lee2013pseudo}. 

\subsection{Data Poisoning in Semi-Supervised Learning}
While the focus of data poisoning work to date has been on supervised learning, there is recent work focused on the impact of data poisoning attacks on semi-supervised learning. Poisoning attacks on labeled samples have been developed which target graph-based semi-supervised learning methods by focusing on poisoning labeled samples that have the greatest influence on the inferred labels of unlabeled samples~\citep{liu2019unified, franci2022influence}. \citet{carlini2021poisoning} introduced a poisoning attack on the unlabeled samples which exploits the pseudolabeling mechanism. This is an instance-targeted attack which aims to propagate the target label from confident target class samples to the target samples (from a non-target class) using interpolated samples between them. \citet{feng2022unlabeled} poisons unlabeled samples using a network that transform samples so they appear to the user's network like the target class. Unlike the the traditional goal of backdoor attacks of introducing a backdoor associated with static triggers, they adapt the decision boundary to be susceptible to future transformed samples. 

\citet{yan2021deep} investigate perturbation-based attacks on unlabeled samples in semi-supervised learning similar to us, but find a simple perturbation-based attack has low attack success. Rather they suggest an attack (called DeHiB) that utilizes a combination of targeted adversarial perturbations and contrastive data poisoning to achieve high attack success. We show settings in which simple perturbation-based attacks are highly successful. Additionally, in Section~\ref{sec:gen_attack_framework}, we discuss how our generalized attack framework encompasses the targeted adversarial perturbations used in DeHiB. \citet{shejwalkar2022perils} has work concurrent to ours which also examines the vulnerability of semi-supervised learning to simple backdoor attacks with augmentation-robust triggers. As with our work, they also identify the requirement for augmentation-robust triggers. They suggest a static trigger pattern that is used in successful attacks. While their work focuses on defining the most effective trigger in the absence of other image modifications, our work focuses on how modifications of the poisoned samples (like adversarial perturbations and interpolations) with an augmentation-robust trigger can influence pseudolabel behavior and vary the effectiveness of attacks.

\section{Backdoor Attacks in the Context of Semi-Supervised Learning}

\subsection{Attack Threat Model}

We consider a setting in which a user has a small amount of labeled data $\mathcal{X}$ for training a classification model. This limited labeled data is not enough to achieve the user's desired classification accuracy, so they collect a large amount of unlabeled data $\mathcal{U}$ from less trusted sources and train their model using the FixMatch semi-supervised learning method~\citep{sohn2020fixmatch} to improve accuracy. The adversary introduces poisoned samples $\mathcal{U}_p$ into the unlabeled dataset with the goal of creating  a strong backdoor in the trained network, resulting in samples being classified as a chosen target class $y_t$ when a trigger is present. To evade detection, the adversary tries to introduce this backdoor as soon as possible in training and maintain a high classification accuracy in the model trained with the poisoned samples. Because the poisoned samples are only included in the unlabeled portion of the training data, the adversary can only control the image content for the poisoned samples and not the training labels. The adversary does not have access to the user's network architecture, making this a black-box attack. 

\subsection{FixMatch Details}

FixMatch achieves high classification accuracy with very few labeled samples. It is important to understand details of FixMatch (and similar methods) when aiming to evaluate its potential vulnerability to backdoor attacks. During training, the user has $N_\ell$ labeled samples $\mathcal{X} = \left\{\bm{x_i} : i \in \left(1,...,N_\ell\right)\right\}$ and $N_u$ unlabeled samples  $\mathcal{U} = \left\{\bm{u_i} : i \in \left(1,...,N_u\right)\right\}$. The supervised loss term is the standard cross-entropy loss on the labeled samples. The unique characteristics of FixMatch are incorporated in the unsupervised loss term which utilizes pseudolabeling and consistency regularization. FixMatch approximates supervised learning by estimating pseudolabels $\bm{y^*}$ for the unlabeled samples:
\begin{equation}
    \bm{y^*} = \text{argmax}(f_\theta(T_w(\bm{u}))),
\end{equation}
where $f_\theta(\cdot)$ is the network being trained and $T_w(\cdot)$ is a function that applies ``weak'' augmentations, like horizontal flipping and random cropping, to the samples.

If the confidence of the estimated label is above a user-specified threshold $\tau$, the pseudolabel is retained and used for computing the unsupervised loss term. We define $m_{i}$ as the indicator of which confident pseudolabels are retained: $m_{i} = \1\left(\text{max}(f_\theta(T_w(\bm{u}_i))) > \tau\right)$. The unsupervised loss term is a consistency regularization term which encourages the network output of a strongly augmented sample to be the same as the pseudolabel estimated from the associated weakly augmented sample:
\begin{equation}
    \ell_u = \frac{1}{\sum m_{i}}\sum_{i=1}^{\mu B} m_{i} H(\bm{y^*},f_\theta(T_s(\bm{u}_i))),
\end{equation}
where $B$ is the batch size, $\mu$ is FixMatch unlabeled sample ratio, $H$ is a cross-entropy loss and $T_s(\cdot)$ is a function that applies ``strong'' augmentations like RandAugment~\citep{cubuk2020randaugment}. 

\subsection{Backdoor Attack Vulnerability Considerations}\label{sec:backdoor_vulnerability}
With the consistency regularization and pseudolabeling in mind, we rethink how poisoned samples in backdoor attacks may interact differently in semi-supervised training than in supervised training. 

\textbf{Augmentation-Robust Triggers} Most backdoor attacks have been analyzed in the absence of data augmentations to focus on the impact of the attack itself without introducing augmentation as a confounding factor. However, prior experiments have shown that data augmentation during training can significantly reduce the attack success rates~\citep{li2020rethinking,schwarzschild2021just}. Therefore, to understand the potential effectiveness of backdoor attacks against FixMatch, it is important to use a trigger that is robust to both the weak and strong augmentations that are crucial to its success. We prioritize the robustness of the triggers to data augmentation over their conspicuousness in order to understand the worst case attack potential before focusing on trigger imperceptibility.

\textbf{Estimating Poisoned Labels} In backdoor attacks on supervised learning, the adversary can fix a training label for every poisoned sample and apply triggers to samples that are confusing given these training labels. This forces the network to rely on the trigger to effectively classify poisoned samples as their poisoned training labels. In attacks on the unlabeled data in semi-supervised learning, the adversary is unable to specify training labels and instead the network is responsible for estimating pseudolabels during training. This reliance on the pseudolabels of poisoned samples adds new considerations when understanding backdoor attacks. First, the adversary can try to control the expected pseudolabels through the image content itself. Second, because the pseudolabels are estimated using the current network state, the training labels assigned to poisoned samples will vary during training as the network is updated. Finally, only poisoned samples with confident network outputs will impact the network updates. We suggest that attacks against semi-supervised learning be developed and understood by considering how an adversary may vary the image content in a way that influences the expected pseudolabel outputs.

\textbf{Pseudolabel-Influencing Attacks} To analyze the impact of pseudolabel behavior on attack success, we use two types of attacks. The first attack type, inspired by clean label backdoor attacks in supervised learning~\citet{turner2019label}, uses untargeted adversarial perturbations to influence estimated network outputs. Adversarial perturbations are optimized to achieve misclassification of the images while constraining perturbation magnitude. These perturbation-based attacks can vary from having no perturbations (i.e., the original training images with triggers added) to having large perturbations that significantly vary the image appearance. 

The second type of attack uses poisoned images that are interpolated between target class samples and non-target class samples. Each poisoned sample $\bm{u^*_i}$ is defined as $\bm{u^*_i} = \left(1-\alpha\right)\bm{u^t_i} + \alpha\bm{u^n_i}$, where $\bm{u^t_i}$ is a sample from the target class, $\bm{u^n_i}$ is a sample from a non-target class, and $\alpha \in \left[0,1\right]$ defines the interpolation ratio. By varying the contribution from each sample through $\alpha$, we can make samples more or less likely to result in target class pseudolabels. Notably ~\citet{carlini2021poisoning} uses image interpolation to define an instance-targeted attack against semi-supervised learning. Both of us use interpolated samples to influence expected pseudolabel behavior. Carlini aims to transfer a target pseudolabel from one sample to another through a path made up of interpolated samples. He does this to adapt the decision boundary around a target sample. We aim to use interpolation to influence the expected pseudolabel outputs of all of the poisoned samples. Therefore, Carlini uses interpolation to narrowly target individual sample instances and we use interpolation to more globally encourage certain behaviors from the poisoned samples. 

\begin{figure}
     \centering
     \begin{subfigure}[b]{0.47\textwidth}
         \centering
         \includegraphics[width=\textwidth]{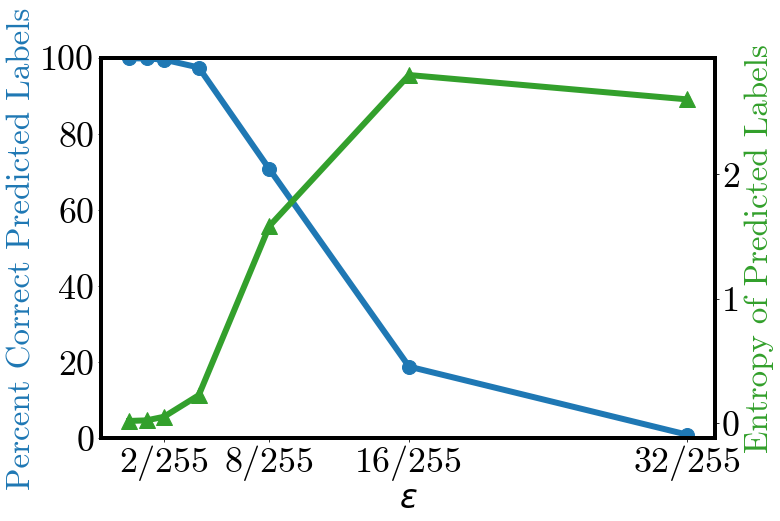}
         \caption{Perturbation-Based Attacks}
         \label{subfig:pseudoEnt}
     \end{subfigure}
     \begin{subfigure}[b]{0.47\textwidth}
         \centering
         \includegraphics[width=\textwidth]{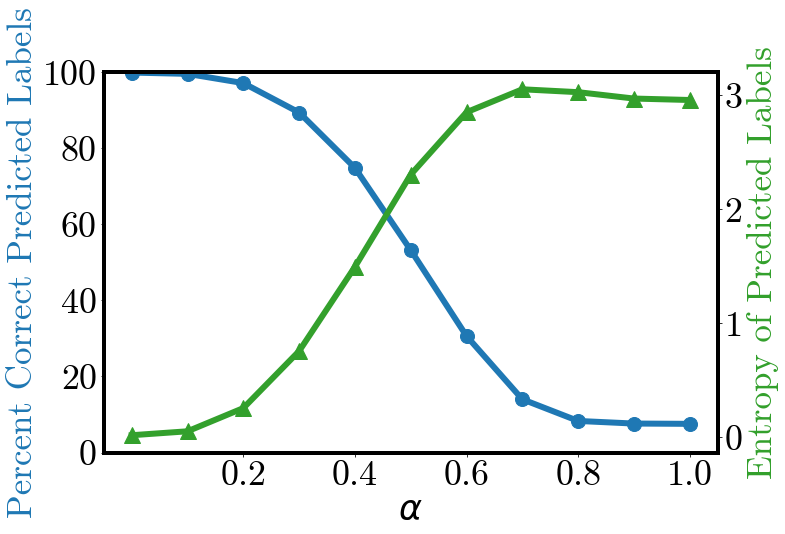}
         \caption{Interpolation-Based Attacks}
         \label{subfig:pseudoEnt_interp}
     \end{subfigure}
        \caption{Predicted labels of modified samples. (a-b) Percentage of modified training samples with the ground truth class as the estimated label (blue circle line) and the entropy of the distribution of predicted labels (green triangle line) as modification strength is varied. (a) Impact of perturbation-based attacks as $\epsilon$ is varied. (b) Impact of interpolation-based attacks as $\alpha$ is varied.}
        \label{fig:pseudoSpecturm}
\end{figure}

Both of these image modifications, perturbation and interpolation, have the potential to influence the behavior of predicted class outputs. To understand how the strength of the modification impacts the distribution of estimated network outputs, we examine the outputs from a network trained using supervised learning on CIFAR-10 training samples. For perturbation-based attacks, we use Projected Gradient Descent (PGD) adversarial perturbations~\citep{madry2018towards}, varying the constraint $\epsilon$ on the $\ell_\infty$ norm of the perturbation magnitude. For interpolation-based attacks, we interpolate between our selected target class samples and randomly selected non-target class samples while varying $\alpha$. We apply triggers and weak augmentations to the images to model the poisoned samples in semi-supervised learning. Fig.~\ref{fig:pseudoSpecturm} shows the impact of perturbation and interpolation strength on pseudolabel outputs. The blue line in each plot is the average percentage of modified samples with estimated network outputs that match their ground truth class and the green line is the average entropy of the distribution of class outputs for modified samples. As the perturbation strength increases, fewer poisoned samples are estimated to be the ground truth label and the entropy of the distribution of network outputs increases, indicating the class estimates are distributed more evenly across all class outputs. The same pattern of behavior is seen with the interpolation-based attacks, suggesting similar attack performance between perturbation-based attacks and interpolation-based attacks. While this test is run against a fully trained network, it gives us useful insights for reasoning about the pseudolabels during semi-supervised learning. At low perturbation strengths and small $\alpha$ interpolation values, we expect most poisoned samples have their ground truth classes as pseudolabels. At greater perturbation strength and larger $\alpha$ values, we expect most poisoned samples will not have their ground truth classes as pseudolabels and instead their pseudolabels will be relatively evenly distributed across other classes. 

The results above suggest that both perturbation-based attacks and interpolation-based attacks are successful at impacting pseudolabel behavior. We begin with perturbation-based attacks because they have been proven successful in clean label backdoor attacks against supervised learning in which the training labels constrain the network output. By implementing perturbation-based attacks against semi-supervised learning without training labels, we can directly compare to prior attacks performance on supervised learning. However, with our threat model in which there are limited labeled samples, the requirement for an adversary to obtain enough labeled data to fully train a network for generating adversarial attacks is unrealistic in practical settings. Interpolation-based attacks present a simpler and more realistic attack alternative to perturbation-based attacks. Adversaries would only require a limited number of samples from target and non-target classes to generate interpolation-based poisoned samples. Therefore, we will continue our investigation with interpolation-based attacks which could be implemented by a much less sophisticated adversary with limited data. 

\section{Analysis}\label{sec:analysis}

We begin our analysis of the vulnerability of semi-supervised learning methods to our pseudolabel-influencing attacks by considering the following experimental setup.

\textbf{Datasets} We generate attacks using the CIFAR-10 dataset~\citep{krizhevsky2009learning} with 50,000 training images and 10,000 test images from 10 classes. We chose this dataset because it is a standard benchmark dataset used for studying both semi-supervised learning and data poisoning. 

\textbf{Semi-Supervised Learning Methods} We perform our analysis on FixMatch~\citep{sohn2020fixmatch} which achieves a classification accuracy of $94.93\%$ on CIFAR-10 with only 250 labeled samples. We largely follow the experimental details from~\citep{sohn2020fixmatch}, using a WideResNet-28-2~\citep{zagoruyko2016wide} architecture, RandAugment~\citep{cubuk2020randaugment} for strong augmentation, and horizontal flipping and cropping for weak augmentation. We experiment with 250 labeled samples. Because we are focused on analyzing the attack dynamics and define a threat model in which the adversary wants to introduce the backdoor as soon as possible during training, we limit each experiment to 100,000 training steps rather than the $2^{20}$ training steps used in the original FixMatch implementation. We found that these shorter training runs achieve relatively high classification accuracy (around $90\%$) and attacks often reach a stable state long before the end of the runs. See Appendix~\ref{app:fixmatch_details} for a detailed description of the FixMatch training implementation.

\textbf{Poisoning Attack} We define the target class of the attack as the ground truth class from which we select samples to be poisoned. In our perturbation-based attacks, similar to clean-label backdoor attacks, we perturb our poisoned samples using adversarially trained ResNet models~\citep{madry2018towards}. Our interpolated poisoned samples are generated as described in Section~\ref{sec:backdoor_vulnerability}. Triggers are added after the images are modified. As discussed in Section~\ref{sec:backdoor_vulnerability}, we begin our analysis using augmentation-robust triggers. In particular, we use the four-corner trigger, suggested in ~\citet{turner2019label} for its invariance to flipping and visibility under random cropping (see Figs.~\ref{fig:poison_img} and~\ref{fig:poison_img_interp} for examples of poisoned images). This trigger is robust to strong augmentations. We define poisoning percentages with respect to the entire training set.

\textbf{Metrics} We analyze two metrics when determining the success of backdoor attacks against semi-supervised learning methods. First is the test accuracy which is the standard classification accuracy computed on the test images. Second is the attack success rate which is the percentage of non-target samples from the test set that are predicted as the target class when triggers are added to them. This indicates the strength of the backdoor in the trained network.

\subsection{Success of Simple Pseudolabel-Influencing Attacks}

\begin{figure}
     \centering
     \begin{subfigure}[b]{0.64\textwidth}
         \centering
         \includegraphics[width=0.98\textwidth]{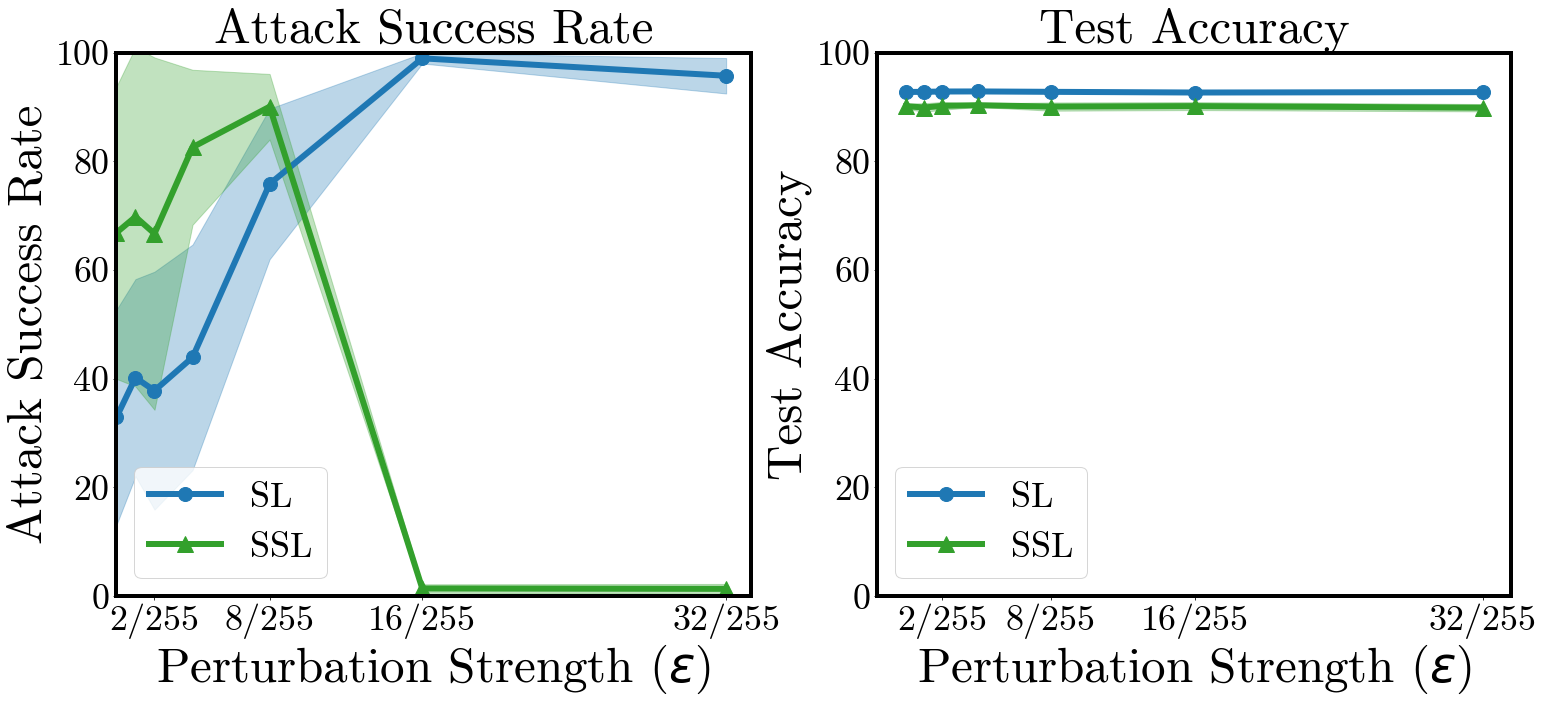}
         \caption{}
         \label{fig:asr_varyeps}
     \end{subfigure}
     \hfill
     \begin{subfigure}[b]{0.33\textwidth}
         \centering
         \includegraphics[width=0.98\textwidth]{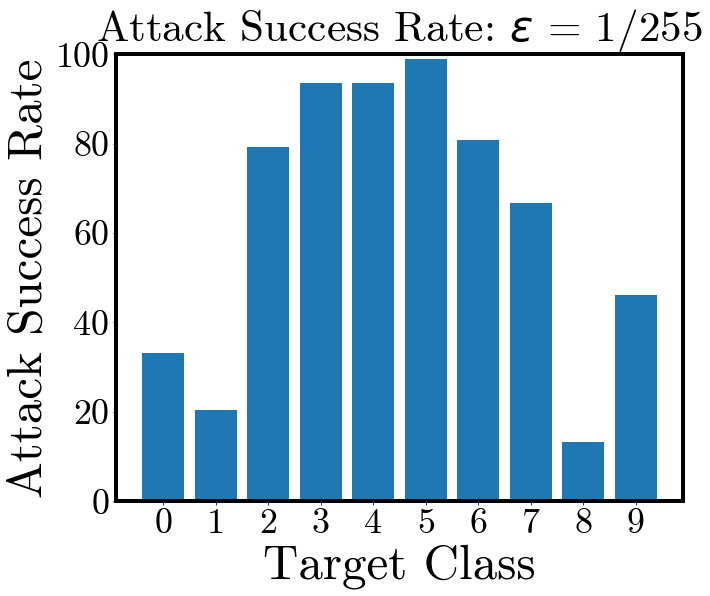}
         \caption{}
         \label{subfig:asr_1_255}
     \end{subfigure}
        \caption{(a) Performance of perturbation-based attacks against supervised learning (blue circle line) and semi-supervised learning (green triangle line) with varying $\epsilon$. (b) Attack success rate from a weak perturbation attack ($\epsilon = 1/255$) as the target class is varied.}
        \label{fig:asr_overview}
\end{figure}

We first examine the performance of simple perturbation-based backdoor attacks as we vary the constraint $\epsilon$ on the magnitude of the adversarial perturbations (see Fig.~\ref{fig:asr_varyeps}). For each $\epsilon$, we run five trials, varying the target class for each run from classes 0-4, and poison $1\%$ of the entire dataset (i.e., 500 target class samples). The poisoned samples are perturbed and have the four corner trigger added. We compare the performance of the attacks against supervised learning (blue line) and semi-supervised learning (green line). Note these perturbation-based attacks against supervised learning, when the adversary sets training labels, are the same as clean-label backdoor attacks~\citep{turner2019label}. The test accuracy is stable as we vary perturbation strength and the resulting accuracy with semi-supervised learning is slightly lower than the accuracy with supervised learning. This is expected because supervised learning uses all the training labels, and we are analyzing the shorter FixMatch training runs which do not reach their maximum test accuracy as detailed above.

These results show several interesting characteristics of the performance of backdoor attacks. First, the attacks against semi-supervised learning are highly successful for moderate perturbation strengths with an average attack success rate of $93.6\%$ for the attacks with $\epsilon = 8/255$ compared to an average attack success rate of $82.58\%$ for the attacks on supervised learning. Second, there is a large variation in the attack success rates for weak perturbations. Fig.~\ref{subfig:asr_1_255} shows the attack success rate for each attack against semi-supervised learning with $\epsilon = 1/255$. While several attacks have very high attack success rates, the attack success rates for the attacks against classes 0, 1, and 8 are low. When comparing against supervised learning, the average attack success rate for weak perturbation attacks is high but the attacks are not consistently effective across target classes. 

\citet{turner2019label} motivated the creation of their clean-label backdoor attacks against supervised learning using the fact that poisoned samples with the ground truth training label and no perturbations resulted in low attack success rates. We confirm this through the relatively low average attack success rate of $32.9\%$ from unperturbed samples ($\epsilon = 0$) against supervised learning. However, the unperturbed attack against semi-supervised learning is surprisingly effective with an average attack success rate of $73.7\%$ while also having the high variance we see with the low-perturbation attacks (see Fig.~\ref{fig:asr_0_255} for the attack success rate per target class). The final notable characteristic is the very low attack success rate for large perturbation attacks. While attack success rates against supervised learning continue to increase with larger perturbations, the attacks fail against semi-supervised learning. In Section~\ref{sec:discussion} we discuss the possible reasons for this attack behavior. 

\begin{figure}
     \centering
         \includegraphics[width=0.6\textwidth]{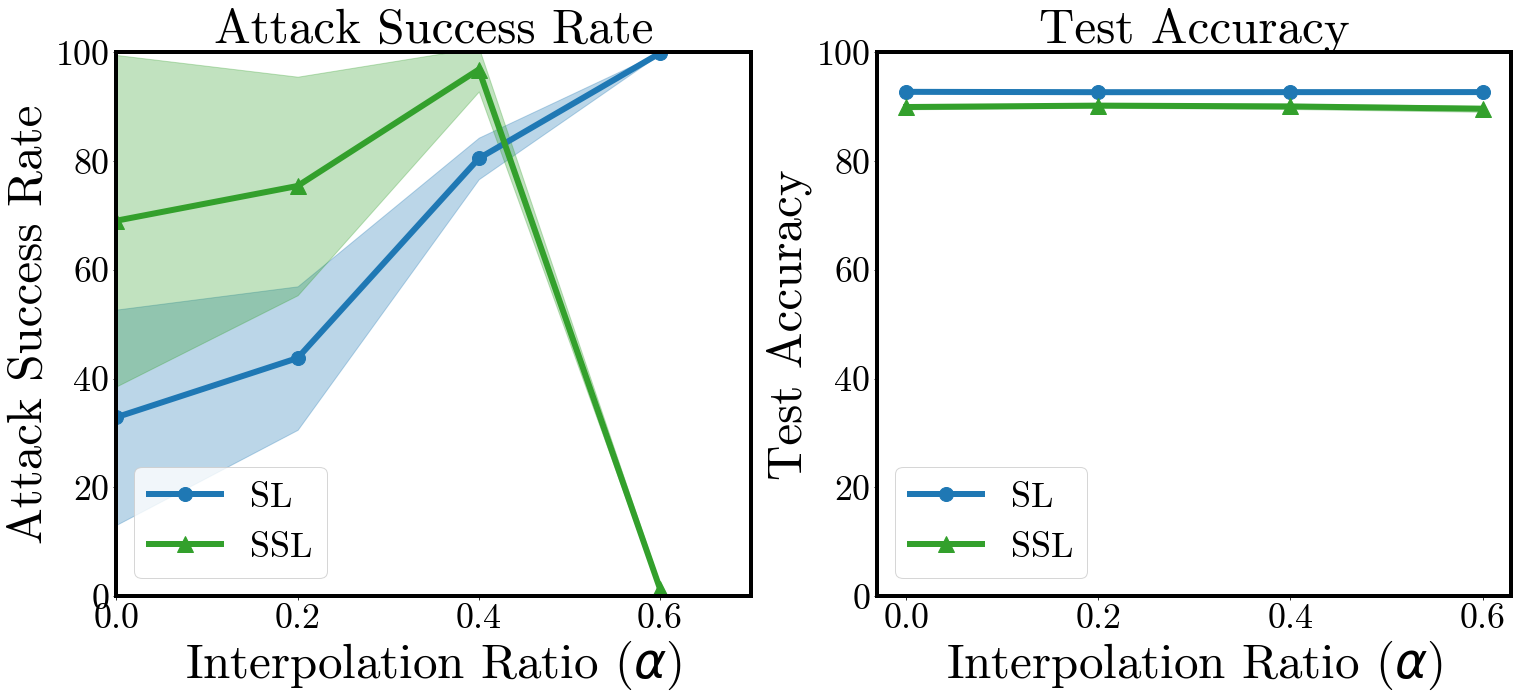}
        \caption{Performance of interpolation-based attacks against supervised learning (blue circle line) and semi-supervised learning (green triangle line) with varying $\alpha$. }
        \label{subfig:asr_varyalpha}
\end{figure}

Fig.~\ref{subfig:asr_varyalpha} shows the performance of the interpolation-based attacks as we vary $\alpha$. Notably, these results show a vary similar pattern to the perturbation-based attacks. At a small interpolation ratio of $\alpha = 0.2$, the average attack success rate is high but there are several target classes with low attack performance, including classes 1 and 8 which also have low attack success rate with weak perturbation-based attacks.  At a moderate interpolation ratio of $\alpha = 0.4$, the attack success rate is consistently high across target classes. Finally, at a large interpolation ratio of $\alpha = 0.6$, the attacks fail and the test accuracy starts to decrease. 

\subsection{Dynamics of Attack Success}\label{sec:attack_dynamics}

\begin{figure}
     \centering
     \begin{subfigure}[b]{0.67\textwidth}
         \centering
         \includegraphics[width=\textwidth]{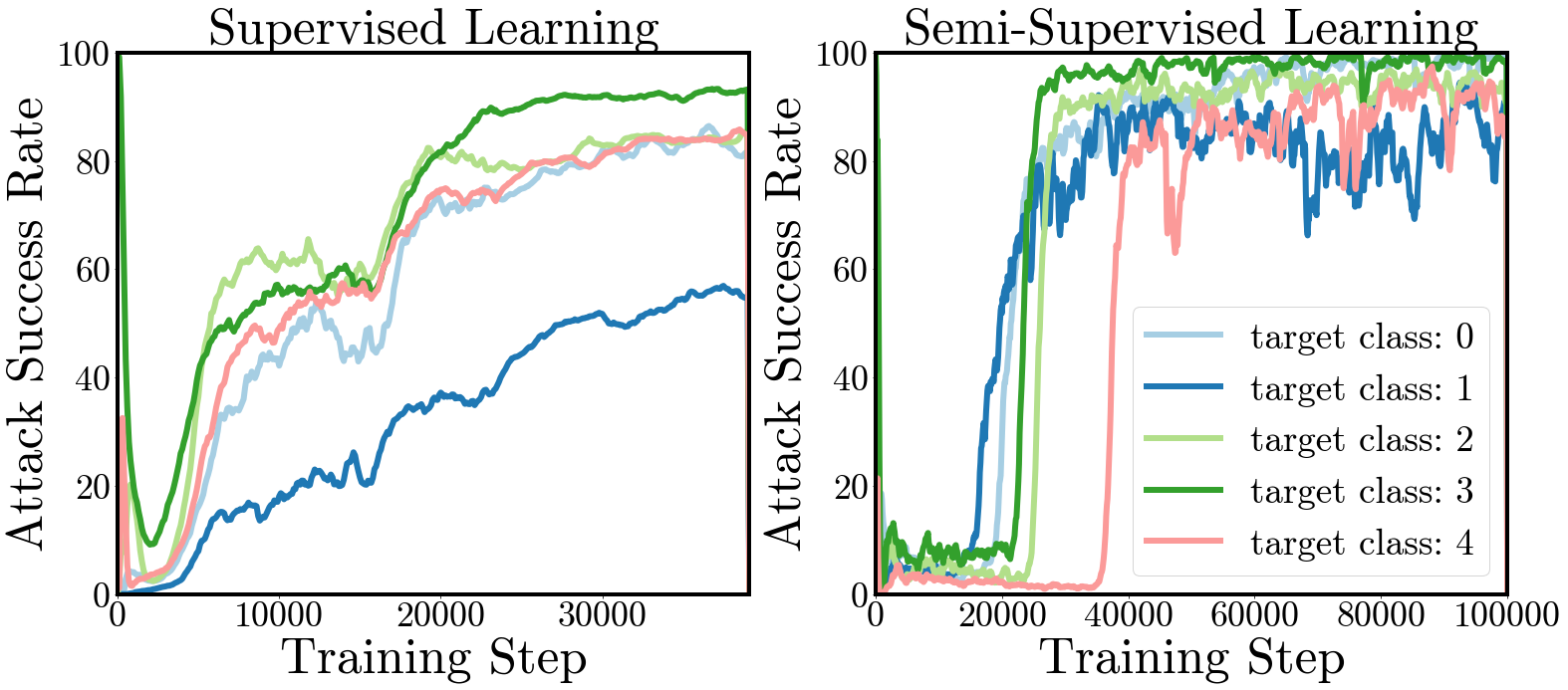}
         \caption{Attack Success Rate Training Dynamics}
         \label{subfig:train_dynamics}
     \end{subfigure}
     \hfill
     \begin{subfigure}[b]{0.98\textwidth}
         \centering
         \includegraphics[width=\textwidth]{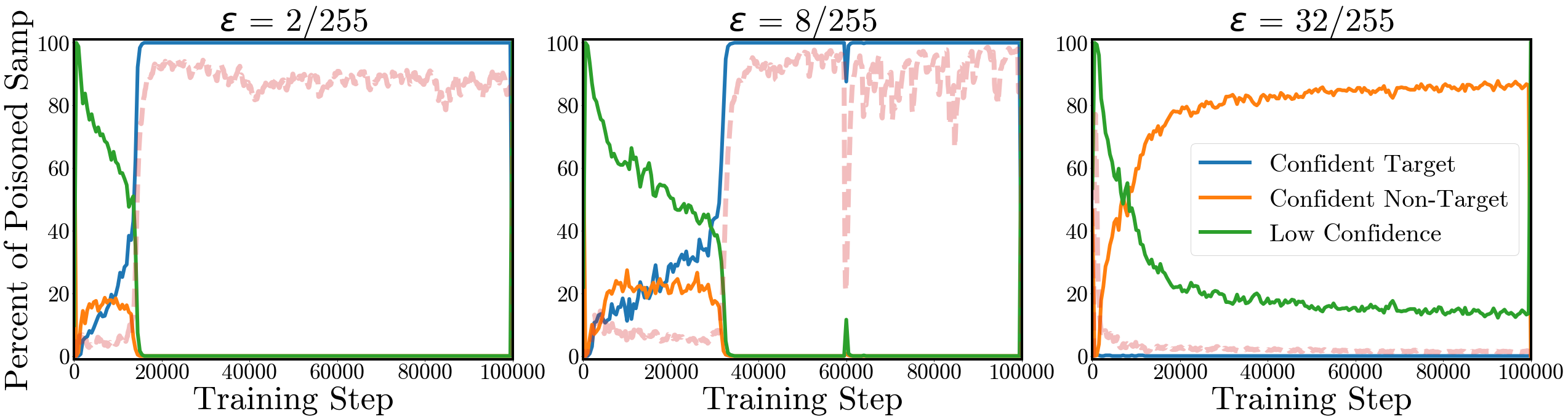}
         \caption{Pseudolabel Behavior}
         \label{subfig:pseudo_behave}
     \end{subfigure}
        \caption{Attack dynamics for perturbation-based attacks. (a) Attack success rate evolution during supervised learning (left) and semi-supervised learning (right). (b) Evolution of the pseudolabel types at various $\epsilon$. The solid lines show the percentage of poisoned samples confident in the target class (blue), confident in a non-target class (orange), or not confident in any class (green). The dashed pink line shows the attack success rate.}
        \label{fig:train_details}
\end{figure}

To understand the dynamics of backdoor attacks against semi-supervised learning, we examine the evolution of the attack success rate during training. Fig.~\ref{subfig:train_dynamics} compares the attack success rates during training between supervised learning and semi-supervised learning with perturbation-based attacks. In supervised learning, which uses a multi-step learning rate scheduler, the attack success rate increases gradually from early in training with jumps at steps down in the learning rate. By contrast, the attack success rate during semi-supervised learning remains low for many training steps until a point in training at which it rapidly increases to a high attack success rate where it remains throughout the rest of training. This suggests that there is a tipping point at which the network forms a backdoor that strengthens rapidly. Fig.~\ref{subfig:pseudo_behave} shows details of the type of pseudolabels the poisoned samples have during training for attacks with weak, moderate, and strong perturbations ($\epsilon = 2/255, 8/255, 32/255$ respectively). The blue lines indicate the percentage of poisoned samples that are confidently estimated as the target class (i.e., the predicted confidence in the target class is above the threshold $\tau$). The orange lines indicate the percentage of poisoned samples that are confidently estimated as a non-target class. The green lines show the percentage of poisoned samples in which the predicted class estimates do not surpass the confidence threshold. The dashed red line is the attack success rate for reference. Of interest are the weak and moderate perturbation attacks in which the percent of poisoned samples with confident target class estimates increases steadily until a point at which nearly all poisoned samples become confident in the target class very rapidly, even if they were previously confident in another class. This suggests that as the backdoor begins to strengthen, it results in poisoned samples which were previously confusing to the network being assigned the target class as a pseudolabel. Similar attack dynamics are observed with interpolation-based attacks (See Appendix~\ref{app:attack_dynamic_interp}) .

\section{Discussion}\label{sec:discussion}

In the previous section, we showed that both simple perturbation and interpolation-based attacks are very successful against semi-supervised learning models. In addition to the results above showing the success of attacks using weak and moderate perturbations and interpolations, Appendix~\ref{app:additional_exp} shows the performance of the perturbation-based attacks with pretrained networks and as we vary the number of labeled samples, the percentage of poisoning, the type of trigger, and the semi-supervised learning technique. In all of these cases, we see that the moderate perturbation attacks with augmentation-robust triggers are highly effective. As we work to understand the reasons for attack success and failure on semi-supervised learning, we recognize that the our two attack types both influence two major factors that impact attack performance: the distribution of estimated pseudolabels and the clarity of class-specific features in the poisoned samples. We reason about the observed attack performance by discussing how the strength of the image modifications (perturbation and interpolation) impact these two factors.

\begin{figure}
     \centering
     \begin{subfigure}[b]{0.45\textwidth}
         \centering
         \includegraphics[width=\textwidth]{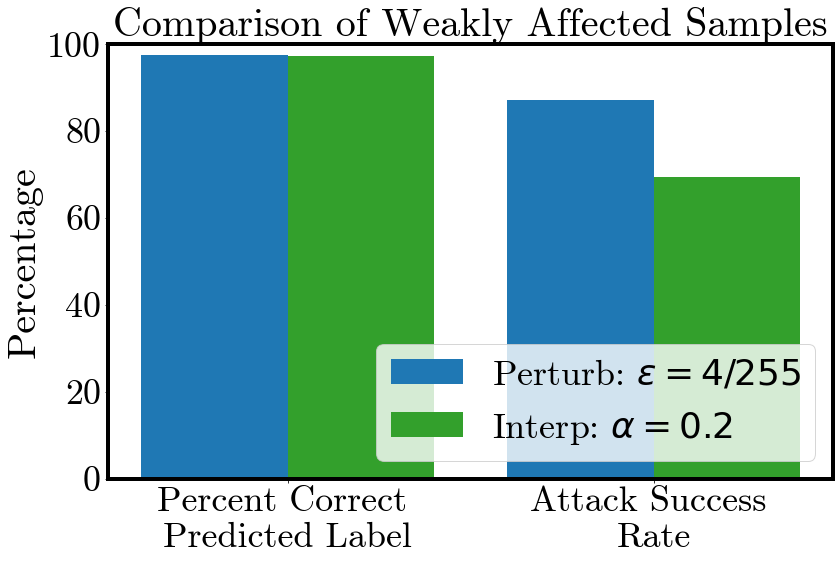}
         \caption{Weakly Modified Samples}
         \label{subfig:comp_weak}
     \end{subfigure}
     \hfill
     \begin{subfigure}[b]{0.47\textwidth}
         \centering
         \includegraphics[width=\textwidth]{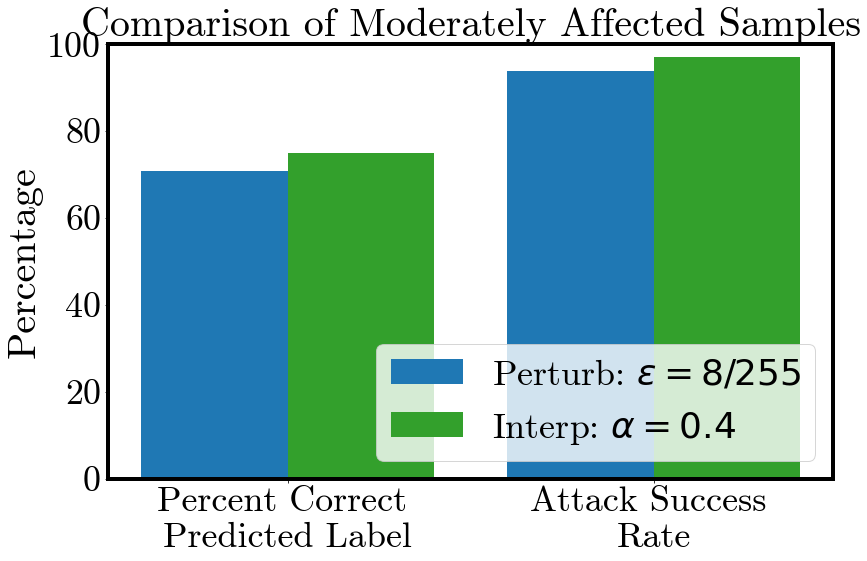}
         \caption{Moderately Modified Samples}
         \label{subfig:comp_moderate}
     \end{subfigure}

    \begin{subfigure}[b]{0.5\textwidth}
         \centering
         \includegraphics[width=\textwidth]{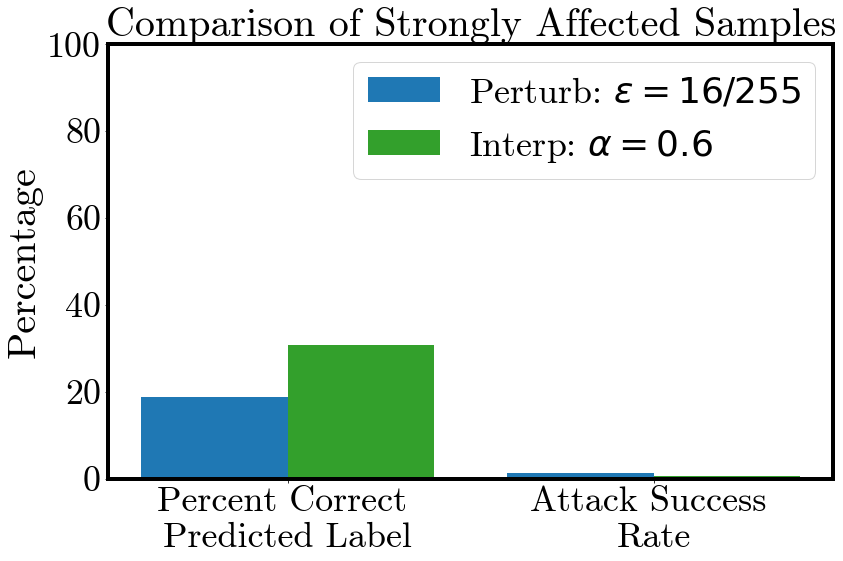}
         \caption{Strongly Modified Samples}
         \label{subfig:comp_strong}
     \end{subfigure}
        \caption{Comparison of perturbation and interpolation-based attacks. In each plot the percent correct predicted label bar indicates the percentage of poisoned samples that are correctly predicted as the target class by a network fully trained in a supervised manner. The attack success rate bar shows the average attack success rate. (a) Weakly modified samples. (b) Moderately modified samples. (c) Strongly modified samples.}
        \label{fig:comp_perturb_interp}
\end{figure}

First we compare the behavior of perturbation-based attacks and interpolation-based attacks. Fig.~\ref{fig:comp_perturb_interp} compares the predicted label behavior and attack success rate for attacks using perturbation and interpolation. In each of these plots the "Percent Correct Predicted Labels" represents the percent of modified samples that are correctly predicted as their ground truth class by a network fully trained using supervised learning. These values are obtained from the experiments plotted in Fig.~\ref{fig:pseudoSpecturm}. Each subplot of Fig.~\ref{fig:comp_perturb_interp} shows the performance of perturbation and interpolation-based attacks with weak, moderate, and strong modifications to the poisoned samples. From these results we see that perturbation and interpolation-based attacks that have a similar percentage of correctly predicted labels have similar attack success rates. This suggests that the predicted label behavior has a significant influence on the success of backdoor attacks against semi-supervised learning. 

When the image modifications are weak or nonexistent, most poisoned samples will receive confident pseudolabels corresponding to the ground truth class label. The poisoned samples will have triggers but they will also have clear target-class-specific features that the network can use for classification, giving the network little reason to rely on the triggers. Notably, even in attacks against semi-supervised learning with weak image modifications, we are seeing high attack success rates for several target classes. However, weak modification attacks against some target classes, like classes 1 (automobile) and 8 (ship) shown in Fig.~\ref{subfig:asr_1_255}, result in weak backdoors. This may indicate that some classes have more distinct features that the network can rely on more strongly, weakening the backdoor. The clean label backdoor attack against supervised learning encourages additional reliance on the trigger by increasing perturbation strength while fixing the training label as the ground truth class, making the samples more difficult to classify. Employing the same technique of increasing perturbation strength in the hope of improving attack performance against semi-supervised learning comes with the additional complication of the perturbations leading to different pseudolabel outputs. 

We see the impact of this complication in the strong modification tests in which most of the samples have pseudolabels that are confident in non-target classes, as seen in the plot of $\epsilon = 32/255$ from Fig.~\ref{subfig:pseudo_behave}. Because the perturbations are untargeted, strong perturbations result in high entropy predicted pseudolabels distributed across many classes, as we see in Fig.~\ref{subfig:pseudoEnt}. In interpolation-based attacks, the high entropy of pseudolabels at large $\alpha$ values results from selecting the non-target interpolation end points $\bm{u_i^n}$ from randomly distributed ground truth classes. In both of these attacks with strong modifications, the network sees samples containing triggers associated with several different classes, leading the network to ignore the trigger as a nuisance feature that does not aid in classification. This shows us how the dependence of semi-supervised learning on pseudolabels limits the effectiveness of attacks at modification strengths that cause too many confident non-target class pseudolabels.

Moderate modification strength attacks are a middle ground in which many poisoned samples will receive confident target class pseudolabels but several samples will be confidently classified as a non-target class or be confusing to the network (the orange and green lines in Fig.~\ref{subfig:pseudo_behave}). These confusing samples will encourage the network to rely more heavily on the triggers, strengthening the backdoor (as seen in the high attack success rate for $\epsilon = 8/255$ attacks in Fig.~\ref{fig:asr_varyeps} and $\alpha = 0.4$ attacks in Fig.~\ref{subfig:asr_varyalpha}).

This analysis suggests that consistently successful backdoor attacks require poison samples that have a pseudolabel distribution heavily concentrated on one class, which can form a weak backdoor, as well as a subset of poisoned samples that are confusing to the network, which can strengthen the backdoor. Next we discuss a generalized attack framework which moves beyond the attacks we have presented to more broadly understand the necessary components for attack success and what leads to attack failure. 

\subsection{Generalized Attack Framework} \label{sec:gen_attack_framework}

Until now we have been analyzing attacks in which all the samples have the same modification strength. This directly links the likely pseudolabel distribution with the difficulty for a network to classify samples. As the modification strength increases, the samples become harder to the network to classify (encouraging a strong backdoor) but the entropy of the pseudolabel distribution also increases (encouraging the network to ignore the trigger). We decouple these two factors using a generalized attack framework which defines attacks that are composed of samples that can be used to create a weak backdoor $U_{pw}$ and samples that are used to strengthen the backdoor $U_{ps}$. The portion of samples from each of these categories is defined by $\lambda$: $N_p = \lambda \lvert U_{pw} \rvert + (1-\lambda)\lvert U_{ps} \rvert$. Weak backdoor-creating samples should be designed to have the same pseudolabel which will be the target class. These samples can be unmodified samples, weakly perturbed samples, weakly interpolated samples, or samples perturbed with strong, targeted adversarial perturbations that are expected to have confident target class pseudolabels. Backdoor-strengthening samples should be confusing to the network and they should initially have low confidence pseudolabels or confident non-target pseudolabels. These samples can be strongly perturbed samples, strongly interpolated samples, unperturbed samples from a class other than the target class, or noisy samples. 

\begin{figure}
     \centering
     \begin{subfigure}[b]{0.46\textwidth}
         \centering
         \includegraphics[width=0.9\textwidth]{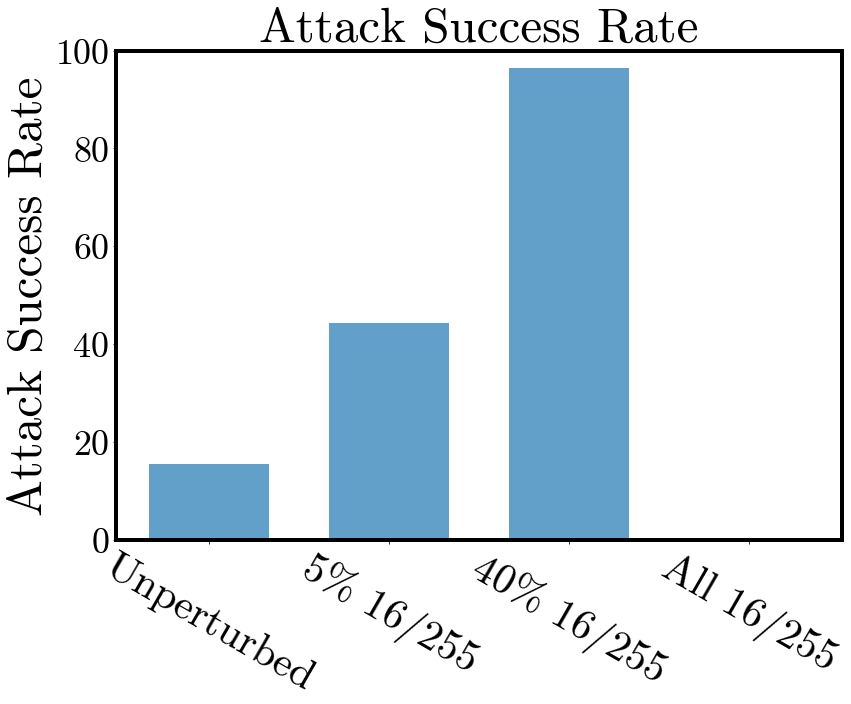}
         \caption{}
         \label{subfig:gen_attack_unperturb}
     \end{subfigure}
     \hfill
     \begin{subfigure}[b]{0.46\textwidth}
         \centering
         \includegraphics[width=0.83\textwidth]{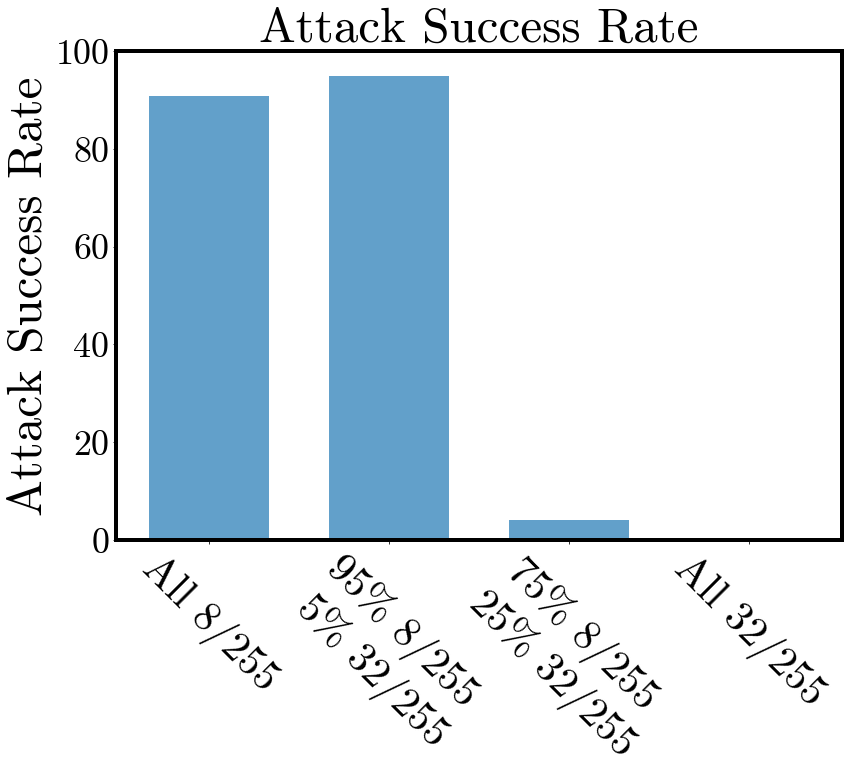}
         \caption{}
         \label{subfig:gen_attack_eps8}
     \end{subfigure}
        \caption{Generalized attack performance with perturbed samples. (a) $U_{pw}$ are unperturbed samples from the target class, $U_{ps}$ are perturbed samples with $\epsilon = 16/255$ for $\lambda = 1, 0.95, 0.6, 0$ (b) $U_{pw}$ are perturbed samples with $\epsilon = 8/255$, $U_{ps}$  are perturbed samples with $\epsilon = 32/255$ for $\lambda = 1, 0.95, 0.75, 0$.}
        \label{fig:gen_attack_results}
\end{figure}

We use this generalized attack framework to generate perturbation-based attacks targeting the automobile class (class 1) with $1\%$ poisoning (i.e., 500 poisoned samples) with results shown in Fig.~\ref{fig:gen_attack_results}. Fig.~\ref{subfig:gen_attack_unperturb} shows attacks in which $U_{pw}$ contains unperturbed samples and $U_{ps}$ contains samples perturbed with $\epsilon = 16/255$. As $\lambda$ is decreased from 1 to 0.95, 0.4 and 0, the attack first becomes more successful with the addition of backdoor strengthening samples. However, too many backdoor strengthening samples causes the attack to fail. Fig.~\ref{subfig:gen_attack_eps8} shows attacks in which $U_{pw}$ contains perturbed samples with $\epsilon = 8/255$ and $U_{ps}$ contains samples perturbed with $\epsilon = 32/255$. At $\lambda = 0.95$, the attack becomes slightly more effective through the addition of only 25 strongly perturbed samples. However, introducing more strongly perturbed samples ($\lambda = 0.75$) leads to attack failure. These results highlight the benefits of the generalized attack framework - varying $\lambda$ can make ineffective attacks more successful, make already successful attacks more successful, and make successful attacks fail.

\begin{figure}
     \centering
     \begin{subfigure}[b]{0.46\textwidth}
         \centering
         \includegraphics[width=0.63\textwidth]{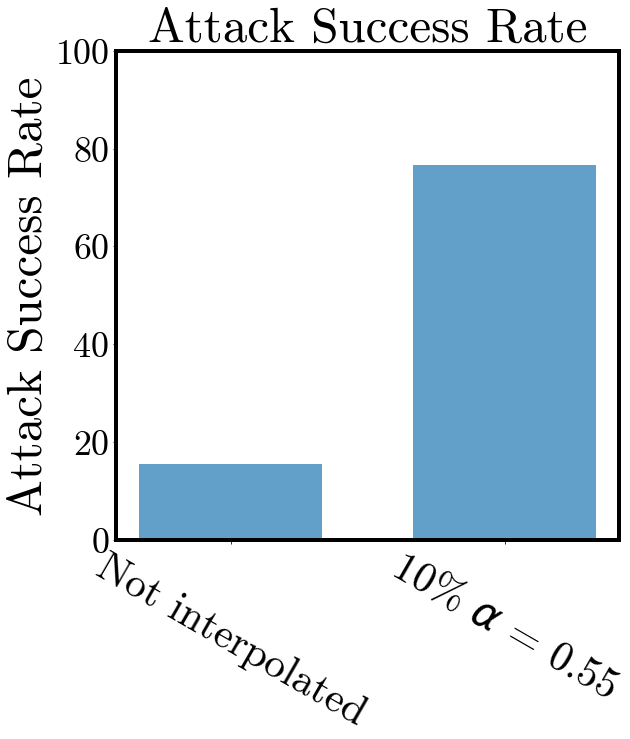}
         \caption{}
         \label{subfig:gen_attack_uninterp}
     \end{subfigure}
     \hfill
     \begin{subfigure}[b]{0.5\textwidth}
         \centering
         \includegraphics[width=0.8\textwidth]{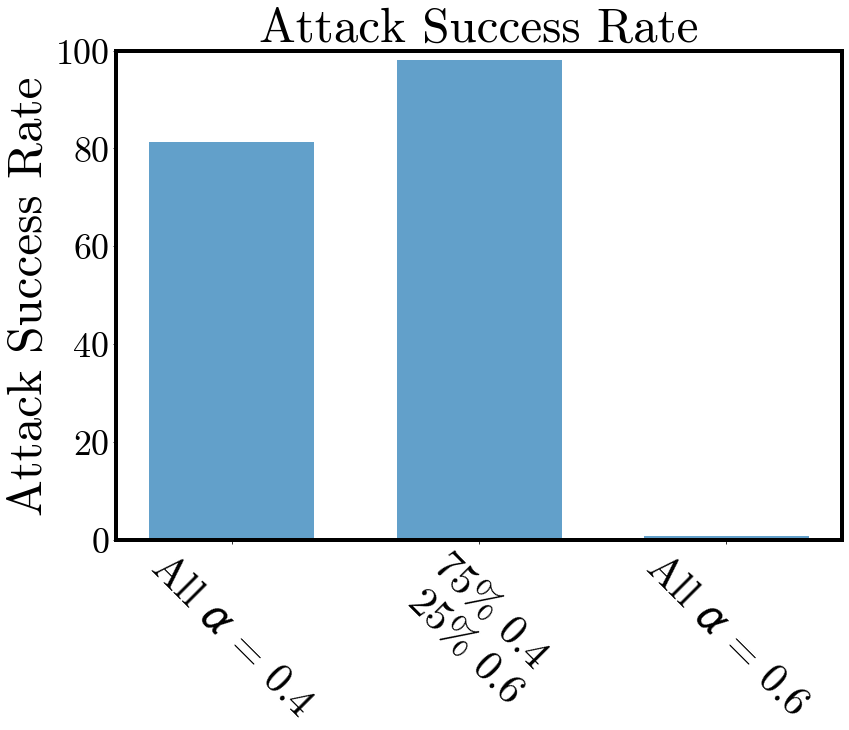}
         \caption{}
         \label{subfig:gen_attack_alpha4}
     \end{subfigure}
        \caption{Generalized attack performance with interpolated samples. (a) $U_{pw}$ are unmodified samples from the target class, $U_{ps}$ are interpolated samples samples with $\alpha = 0.55$ for $\lambda = 1, 0.9$ (b) $U_{pw}$ interpolated samples with $\alpha = 0.4$, $U_{ps}$  are interpolated samples with $\alpha = 0.6$ for $\lambda = 1, 0.75, 0$.}
        \label{fig:gen_attack_results_interp}
\end{figure}

Fig.~\ref{fig:gen_attack_results_interp} shows the performance of the generalized attack framework using interpolation-based attacks targeting the automobile class (class 1) with $1\%$ poisoning (i.e., 500 poisoned samples). In Fig.~\ref{subfig:gen_attack_uninterp}, the attack is defined with unmodified samples in $U_{pw}$ and samples interpolated with $\alpha = 0.55$ in $U_{ps}$. This shows the addition of a only 50 strongly interpolated samples can make an attack significantly more effective. Fig.~\ref{subfig:gen_attack_alpha4} shows attacks in which $U_{pw}$ contains interpolated samples with $\alpha = 0.4$ and $U_{ps}$ contains strongly interpolated samples with $\alpha = 0.6$. As in the perturbation-based generalized attacks above, the addition of a limited number of strongly modified samples can improve an already successful attack but too many strongly modified samples leads to attack failure.

The large variation in attack performance due to relatively small variations in the portion of samples that are confusing to the network suggests a potential focus point for defenses against these types of attacks on semi-supervised learning. The inclusion of a small number of very confusing samples with triggers significantly reduces the impact of the attack.

\subsection{Defenses}

We view our analysis of perturbation and interpolation-based attacks against semi-supervised learning and our introduction of a generalized attack framework as a starting point towards understanding and defending against backdoor attacks targeting semi-supervised learning. We showed that backdoor attacks are very effective against semi-supervised learning in certain settings (i.e., with augmentation-robust triggers and moderate modification strength) but fail in others. This knowledge can be used to define the maximally effective attacks which can be the focus of proposed defenses.

Standard defenses that probe networks after they are trained~\citep{liu2017neural,kolouri2020universal,liu2018fine,liu2019abs,wu2021adversarial} should work similarly on networks trained using both supervised and semi-supervised learning because backdoor attacks have the same goal in both of those cases. Other established defenses focus on cleansing the training data by identifying poisoned samples~\citep{chen2018detecting,tran2018spectral} or reverse-engineering triggers~\citep{wang2019neural,qiao2019defending,guo2019tabor}. Both activation clustering~\citep{chen2018detecting} and the spectral signature defense~\citep{tran2018spectral} identify poisoned samples by estimating clusters likely to include poisoned samples using training labels which are not available in unlabeled data. Defenses that reverse-engineer triggers may more easily identify the conspicuous, augmentation-robust four corner trigger used in our analysis. This motivates future investigation into less conspicuous triggers that are also robust to significant data augmentations.

There are unique characteristics of the attacks against semi-supervised learning that suggest avenues for future defenses. First, the labels assigned to poisoned samples in semi-supervised learning vary during training. As we see in~\ref{subfig:pseudo_behave}, many of the poisoned samples are originally classified with pseudolabels other than the target class. This suggests that there may be an effective defense that eliminates samples that rapidly change their pseudolabel during training, limiting the backdoor strengthening samples from influencing the network. Second, we see in Figs.~\ref{fig:asr_varyeps} and ~\ref{fig:gen_attack_results} that poisoned samples that have confident pseudolabels associated with several classes other than the target class significantly reduce the attack success rate. This suggests further investigation into how these samples impact the attack success and how a defender may use these qualities to create a defense.

\section{Conclusion}

We analyzed the effectiveness of backdoor attacks on unlabeled samples in semi-supervised learning when the adversary has no control over training labels. This setting requires a rethinking of attack development which focuses on the expected distribution of pseudolabels for poisoned samples and the difficulty in recognizing their class-specific features. We showed two simple attacks with that influence pseudolabel behavior were consistently effective against semi-supervised learning with moderate modification strength and augmentation-robust triggers, and  we defined a generalized attack framework which can be used to separately define weak backdoor-generating samples and backdoor-strengthening samples. This work highlights a serious vulnerability of semi-supervised learning to backdoor attacks and suggest unique characteristics of these attacks that could be used for targeting defenses in the future.

\subsubsection*{Acknowledgments}
This research was funded by the U.S. Government. The views and conclusions contained in this document are those of the authors and should not be interpreted as representing the official policies, either expressed or implied, of the U.S. Government. This work was supported by DARPA Guaranteeing AI Robustness Against Deception (GARD).

\bibliography{references}
\bibliographystyle{apalike}

\appendix
\section{FixMatch Training Details}\label{app:fixmatch_details}

For the FixMatch implementation, we closely followed the training set up from \citet{sohn2020fixmatch}. We used a WideResNet-28-2~\citep{zagoruyko2016wide} architecture, RandAugment~\citep{cubuk2020randaugment} for strong augmentation, and horizontal flipping and cropping for weak augmentation. We used an SGD optimizer with momentum of 0.9, a weight decay of $5\times 10^{-4}$, and Nesterov momentum. Like \citet{sohn2020fixmatch}, we used a cosine learning rate decay and quoting from them, we set the ``learning rate to $\eta\text{cos}\left(\frac{7\pi k}{16K}\right)$, where $\eta$ is the initial learning rate, $k$ is the current training step, and $K$ is the total number of training steps.'' We ran 25,000 training epochs and each epoch runs through all the batches of the labeled data. Therefore, with 250 labeled samples, there are four steps per epoch and 100,000 steps total. We report the performance on the exponential moving average of the network parameters. We ensured an even distribution of classes in the labeled data. Additional training parameters are shown in Table~\ref{tab:fixmatch_params}. We found the following public github repository a good guide to implementing FixMatch: \url{https://github.com/kekmodel/FixMatch-pytorch}.

\begin{table}[!htb]
\centering
\caption{Training parameters for FixMatch}
\label{tab:fixmatch_params}
\begin{tabular}{||l||} 
 \hline
 FixMatch Training Parameters \\ 
 \hline
 batch size ($B$): 64   \\ 
 number of epochs: 25000 \\
 initial learning rate ($\eta$): 0.03 \\
 total number of training steps ($K$): $2^{20}$ \\
 poisoning percentage (percentage of entire dataset): $1\%$ (500 samples)\\
 number of labeled samples: 250 \\
 confidence threshold ($\tau$): 0.95 \\
 $\mu$: 7 \\
 $\lambda_u$: 1 \\
 \hline
\end{tabular}
\end{table}

\section{Adversarial Perturbation Details}
For our perturbation-based attacks we used samples that were perturbed using PGD attacks against an adversarially trained network. For $\epsilon = 8,16,32/255$ we used perturbed samples provided by the Madry lab whose access locations are specified here:  \url{https://github.com/MadryLab/label-consistent-backdoor-code/blob/main/setup.sh} 
For $\epsilon = 1,2,4/255$ we used perturbed samples generated against a adversarially trained network. The adversarially trained network was a ResNet-50 using $\epsilon = 8/255$ for an $\ell_\infty$ norm. We obtained the weights for the network from the Madry lab: \url{https://github.com/MadryLab/robustness/#pretrained-models}.

\section{Poisoned Sample Details}\label{app:poison_sample}

\begin{figure}
    \centering
    \includegraphics[width=0.98\textwidth]{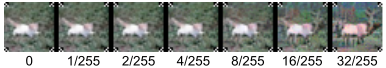}
    \caption{Poisoned images with increasing perturbation strength $\epsilon$ and the four corner trigger.}
    \label{fig:poison_img}

\end{figure}

\begin{figure}
    \centering
    \includegraphics[width=0.5\textwidth]{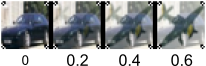}
    \caption{Poisoned images with increasing interpolation ratio $\alpha$ and the four corner trigger.}
    \label{fig:poison_img_interp}

\end{figure}

We used the four corner trigger suggested in ~\citet{turner2019label} , following the example from \url{https://github.com/MadryLab/label-consistent-backdoor-code/blob/main/poison_attack.py}, for creating the attack. Fig.~\ref{fig:poison_img} shows an example of adversarially-perturbed poisoned images with the four corner trigger. Fig.~\ref{fig:poison_img_interp} shows an example of interpolated poisoned images with the four corner trigger.

\section{Supervised Learning Details}\label{app:SL_details}

For supervised learning we also used a WideResNet-28-2 architecture and RandAugment data augmentation during training. We used an SGD optimizer with a momentum of 0.9 and a weight decay of $2\times 10^{-4}$. We used a multi-step learning rate scheduler that reduced the learning rate by $\gamma = 0.1$ at epochs 40 and 60. To stay consistent with our FixMatch experiments, we report the performance on the exponential moving average of the network parameters. 

\begin{table}[!htb]
\centering
\caption{Training parameters for supervised learning}
\label{tab:SL_params}
\begin{tabular}{||l||} 
 \hline
 FixMatch Training Parameters \\ 
 \hline
 batch size: 128   \\ 
 number of epochs: 100 \\
 initial learning rate ($\eta$): 0.1 \\
 poisoning percentage (percentage of entire dataset): $1\%$ (500 samples)\\
 \hline
\end{tabular}
\end{table}

\section{Interpolation-Based Attack Dynamics}\label{app:attack_dynamic_interp}

\begin{figure}
     \centering
     \begin{subfigure}[b]{0.67\textwidth}
         \centering
         \includegraphics[width=\textwidth]{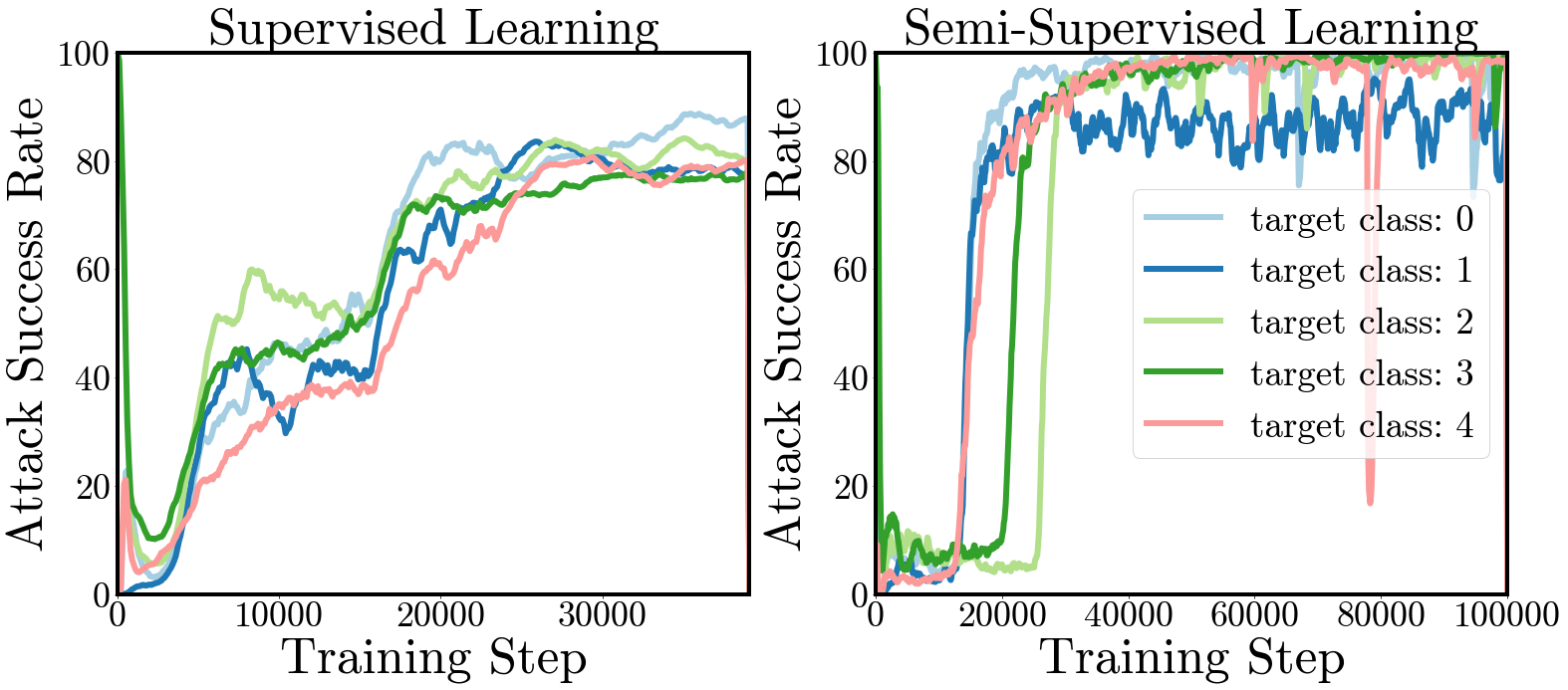}
         \caption{Attack Success Rate Training Dynamics}
         \label{subfig:train_dynamics_interp}
     \end{subfigure}
     \hfill
     \begin{subfigure}[b]{0.98\textwidth}
         \centering
         \includegraphics[width=\textwidth]{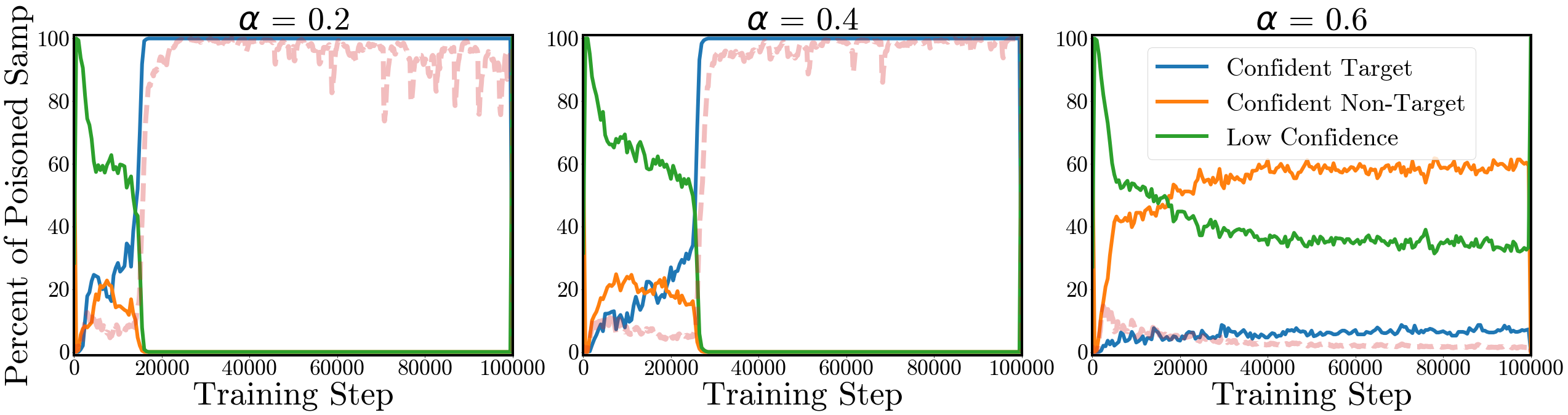}
         \caption{Pseudolabel Behavior}
         \label{subfig:pseudo_behave_interp}
     \end{subfigure}
        \caption{Attack dynamics for interpolation-based attacks. (a) Attack success rate evolution during supervised learning (left) and semi-supervised learning (right). (b) Evolution of the pseudolabel types at various $\alpha$ values. The solid lines show the percentage of poisoned samples confident in the target class (blue), confident in a non-target class (orange), or not confident in any class (green). The dashed pink line shows the attack success rate.}
        \label{fig:train_details_interp}
\end{figure}

Fig.~\ref{subfig:train_dynamics_interp} compares the attack success rates during training between supervised learning and semi-supervised learning with interpolation-based attacks. The training dynamics of these interpolation-based attacks are similar to those described in Section~\ref{sec:attack_dynamics} for perturbation-based attacks. The attacks against supervised learning show a gradual increase in attack success rate whereas the attacks against semi-supervised learning maintain a low attack success rate until a point in training at which the attack success rate increases rapidly. Fig.~\ref{subfig:pseudo_behave_interp} shows details of the type of pseudolabels the poisoned samples have during training for attacks with weak, moderate, and strong interpolations ($\alpha = 0.2, 0.4, 0.6$ respectively). The blue lines indicate the percentage of poisoned samples that are confidently estimated as the target class (i.e., the predicted confidence in the target class is above the threshold $\tau$). The orange lines indicate the percentage of poisoned samples that are confidently estimated as a non-target class. The green lines show the percentage of poisoned samples in which the predicted class estimates do not surpass the confidence threshold. The dashed red line is the attack success rate for reference. The pseudolabel behavior for these interpolation-based poisoned samples is similar to the pseudolabel behavior for perturbation-based samples. One notable difference is, when $\alpha = 0.6$, the percentage of confident non-target pseudolabels is lower than the strong perturbation-based attacks.

\section{Additional Semi-Supervised Learning Experiments}\label{app:additional_exp}

In this section we show results of additional experiments we ran to determine the attack performance varied in different settings. 

\begin{figure}
    \centering
    \includegraphics[width=0.4\textwidth]{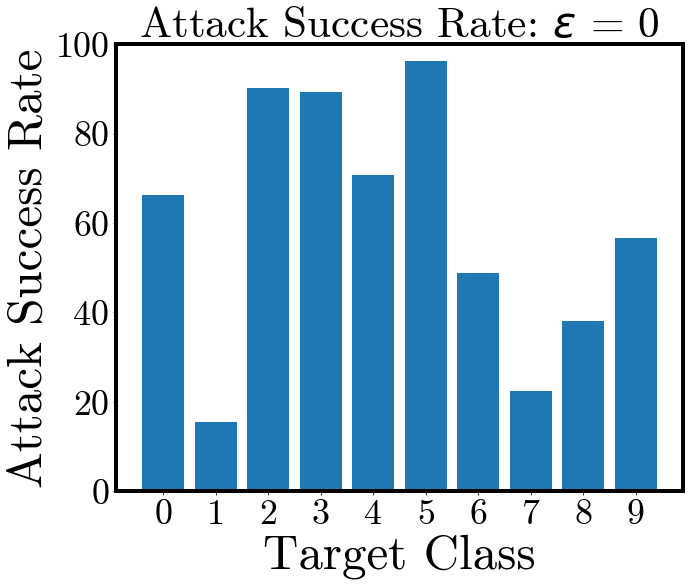}
    \caption{Attack success rate as we vary target class for attacks with unperturbed samples. }
    \label{fig:asr_0_255}
\end{figure}

\begin{figure}
    \centering
    \includegraphics[width=0.7\textwidth]{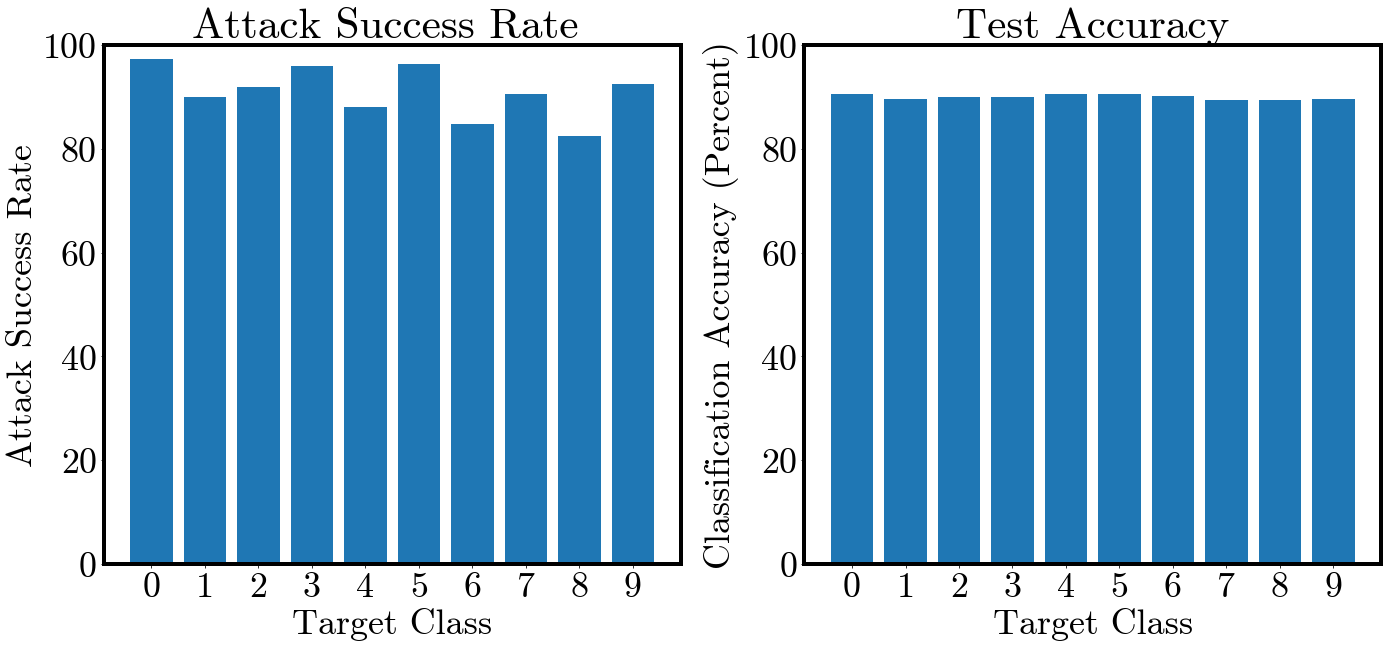}
    \caption{Attack performance as we vary target class for moderate perturbation attacks with $\epsilon = 8/255$ }
    \label{fig:varyTarget}
\end{figure}

\paragraph{Varying Target Class} As we showed in Fig.~\ref{subfig:asr_1_255}, for attacks with weak perturbations, the attack success rate can vary significantly. The attack success rate also varies for attacks that use unperturbed samples, with some attacks achieving very high attack success rates (see Fig.~\ref{fig:asr_0_255}).  However, for attacks with moderate perturbation strength (like $\epsilon = 8/255$) we see fairly consistent attack success rates as we vary the target class (See Fig.~\ref{fig:varyTarget}).

\begin{figure}
    \centering
    \includegraphics[width=0.7\textwidth]{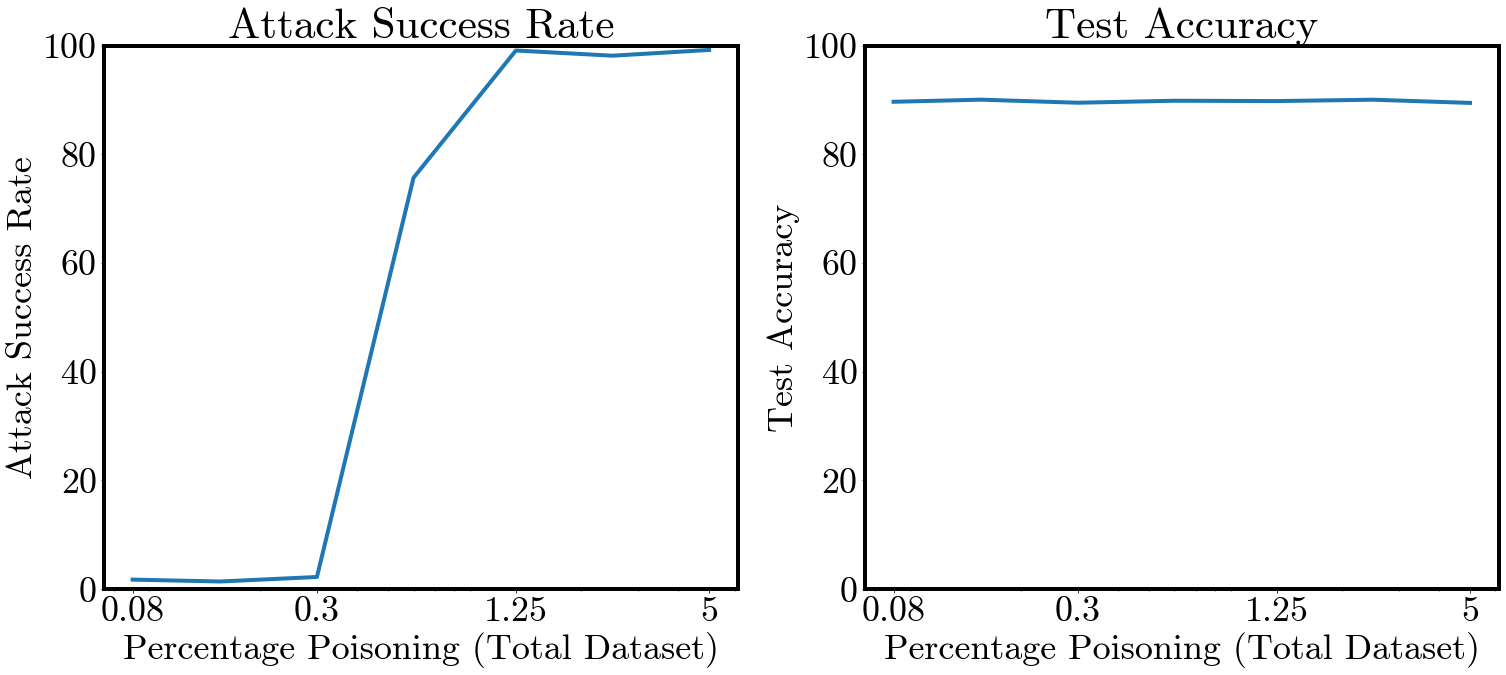}
    \caption{Attack performance as we vary poisoning percentage for perturbation-based attacks with $\epsilon = 8/255$.}
    \label{fig:varyPP}
\end{figure}

\paragraph{Varying Poisoning Percentage} We examine the impact of poisoning percentage on attack performance for moderate perturbation attacks ($\epsilon = 8/255$) in Fig.~\ref{fig:varyPP}. Note that the poisoning percentage is with respect to all 50,000 training samples in the CIFAR-10 dataset. Therefore $0.08\%$ poisoning is 40 poisoned samples and $5\%$ poisoning is 2,500 poisoned samples. The attacks fail for poisoning percentages less than $0.6\%$ after which the attack success rate increases and then plateaus.

\begin{figure}
     \centering
     \begin{subfigure}[b]{0.75\textwidth}
         \centering
         \includegraphics[width=\textwidth]{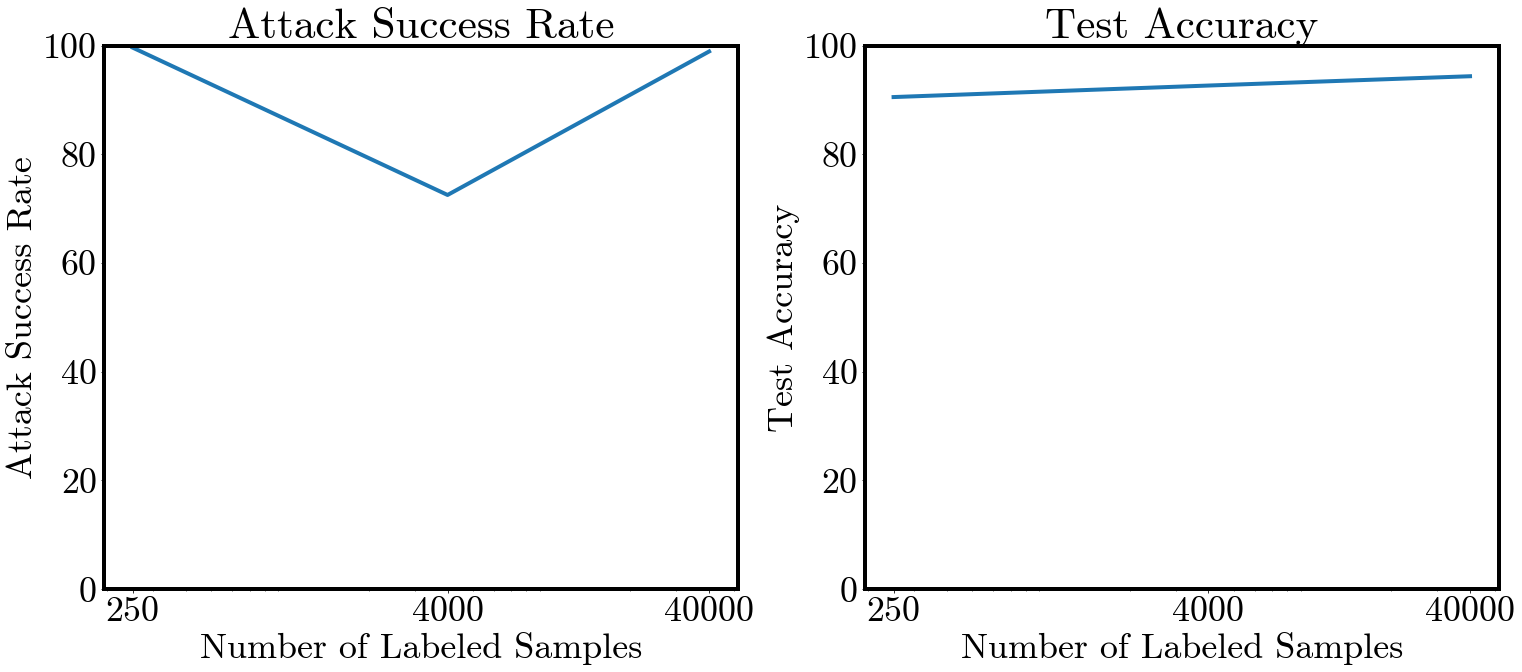}
         \caption{No Pretraining}
         \label{subfig:varyLabeled_nopretrain}
     \end{subfigure}
     \hfill
     \begin{subfigure}[b]{0.75\textwidth}
         \centering
         \includegraphics[width=\textwidth]{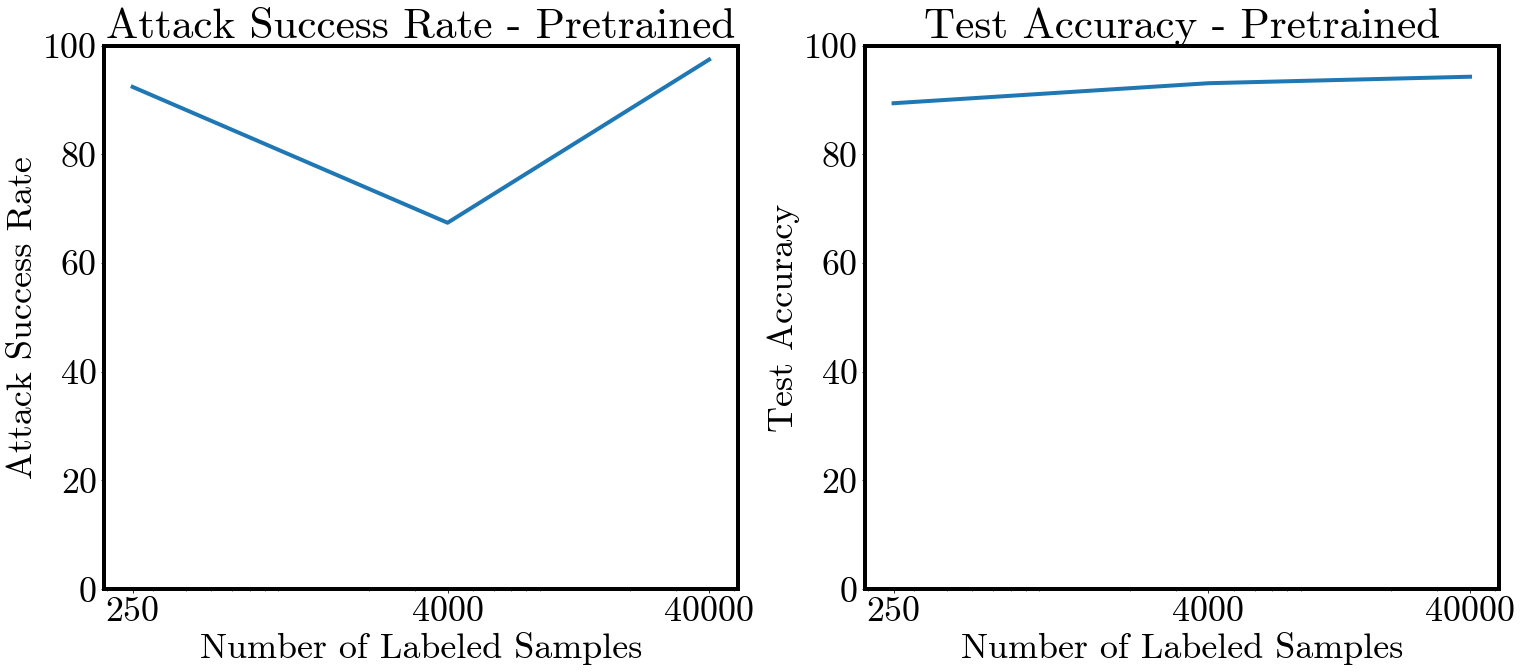}
         \caption{Pretraining}
         \label{subfig:varyLabeled_pretrain}
     \end{subfigure}
        \caption{Attack performance as we vary the number of labeled samples $N_{\ell}$ with and without pretraining.}
        \label{fig:varyPretrain}
\end{figure}

\paragraph{Varying Number of Labeled Samples} We examine the impact of the number of labeled samples both with and without pretraining. Fig.~\ref{subfig:varyLabeled_nopretrain} shows the performance as we vary the number of labeled samples from 250 to 4,000 and 40,000 with $\epsilon$ fixed at $8/255$ and a poisoning percentage of $1\%$. All attacks are successful but the attack with 4,000 labeled samples has a lower attack success rate. Notably these are results for one experiment per $N_\ell$ so there may be natural variations leading to the 4,000 labeled sample run achieving the lowest attack success rate which would be evened out by averaging over multiple runs. Fig.~\ref{subfig:varyLabeled_pretrain} shows the attack performance as we vary the number of labeled samples and perform 20,000 training steps of pretraining with only the labeled samples prior to adding in the unlabeled samples and consistency regularization. The performance looks similar as without pretraining except with slightly lower attack success rates.

\paragraph{Varying the Semi-Supervised Learning Approach} We tested the performance of the perturbation based attack with $\epsilon = 8/255$ against the UDA semi-supervised learning technique~\cite{xie2020unsupervised}. This method is similar to FixMatch in its use of augmentations and consistency regularization. The main difference is that UDA computes the consistency regularization using soft network outputs rather than hard pseudolabels. Table~\ref{tab:varySSL} compares the performance on FixMatch and UDA on target class 0 (airplane). This preliminary experiment confirms other semi-supervised learning methods are likely to be similarly vulnerable to backdoor attacks as FixMatch.

\begin{table}[t]
\caption{Attack performance with varying semi-supervised learning models}
\label{tab:varySSL}
\begin{center}
\begin{tabular}{||c | c | c||} 
 \hline
 Semi-Supervised Learning Technique & Attack Success Rate & Classification Accuracy \\ [0.5ex] 
 \hline\hline
 FixMatch~\citep{sohn2020fixmatch} &  $99.62\%$ & $90.52\%$  \\ 
 \hline
 UDA~\citep{xie2020unsupervised} &  $59.45\%$  &  $92.0\%$ \\
 \hline
\end{tabular}
\end{center}
\end{table}

\paragraph{Vary Trigger Type} We selected the four corner trigger which we found to be robust to strong augmentations and we used this trigger for the experiments presented in this paper. We also tested the effectiveness of single patch triggers in the bottom right of the image (See Table~\ref{tab:varyTrigger}). We see that $8 \times 8$ triggers are also effective against strong augmentations but $4 \times 4$ triggers are not.

\begin{table}[t]
\caption{Attack performance with varying backdoor triggers}
\label{tab:varyTrigger}
\begin{center}
\begin{tabular}{||c | c | c||} 
 \hline
 Trigger Type & Attack Success Rate & Classification Accuracy \\ [0.5ex] 
 \hline\hline
 Four Corner Trigger &  $90.04\%$ & $89.92\%$  \\ 
 \hline
 $8\times 8$ Patch Trigger & $94.50\% $ & $89.36\%$  \\
 \hline
 $4\times 4$ Patch Trigger & $1.37\%$ &  $89.89\%$ \\
 \hline
\end{tabular}
\end{center}
\end{table}

\end{document}